\documentclass{article}

\PassOptionsToPackage{numbers, compress}{natbib}

\usepackage[preprint]{neurips_glfrontiers_2023}

\usepackage[utf8]{inputenc} 
\usepackage[T1]{fontenc}    
\usepackage{hyperref}       
\usepackage{url}            
\usepackage{booktabs}       
\usepackage{amsfonts}       
\usepackage{nicefrac}       
\usepackage{microtype}      
\usepackage[usenames, dvipsnames]{xcolor}         
\usepackage{graphicx}
\usepackage{amsmath}
\usepackage{comment}
\usepackage{floatrow}

\usepackage{floatrow}
\newfloatcommand{capbtabbox}{table}[][0.49\textwidth]

\usepackage{blindtext}

\newtheorem{theorem}{Theorem}
\newtheorem{definition}{Definition}

\title{Higher-Order Expander Graph Propagation}

\author{%
  Thomas Christie \thanks{Equal contribution. Work done when both authors were at the University of Cambridge.}\\
  Unaffiliated\\
  \texttt{thwc3@cantab.ac.uk}
  \And
  Yu He \footnotemark[1]\\
  Stanford University\\
  \texttt{heyu@cs.stanford.edu}
}

\begin{document}

\maketitle

\begin{abstract}
Graph neural networks operate on graph-structured data via exchanging messages along edges. One limitation of this message passing paradigm is the over-squashing problem. Over-squashing occurs when messages from a node's expanded receptive field are compressed into fixed-size vectors, potentially causing information loss. To address this issue, recent works have explored using expander graphs, which are highly-connected sparse graphs with low diameters, to perform message passing. However, current methods on expander graph propagation only consider pair-wise interactions, ignoring higher-order structures in complex data. To explore the benefits of capturing these higher-order correlations while still leveraging expander graphs, we introduce \textit{higher-order expander graph propagation}. We propose two methods for constructing bipartite expanders and evaluate their performance on both synthetic and real-world datasets.
\end{abstract}

\section{Introduction}

Graph neural networks (GNNs) \cite{kipf&welling2017, velickovic2018, gilmer2017neural} have gained significant attention for their effective applications across various domains \cite{do2018chemical, hamaguchi2017knowledge,  wu2019recommendation}. They operate directly on graph-structured data, utilising the inherent symmetries of graphs \cite{bronstein2021geometric}. GNNs commonly employ the message-passing paradigm \cite{gilmer2017neural, velickovic2022message}, where messages are exchanged along the edges of a graph. However, this approach faces some fundamental challenges, including limited expressivity \cite{xu2018how, morris2018neural}, over-smoothing \cite{oversmooth} and over-squashing \cite{alon2020oversquashing}.

In a multi-layered GNN architecture, a node can aggregate information from neighbouring nodes within a certain radius determined by the number of layers, denoted as $k$. The over-squashing problem \cite{alon2020oversquashing} becomes apparent as $k$ increases. In such cases, nodes are compelled to compress information from an exponentially growing number of neighbouring nodes into their fixed-size feature vectors. This compression process can lead to the loss of important information, particularly in long-range interactions between nodes. Addressing this over-squashing issue is crucial for enhancing GNNs' expressivity \cite{digiovanni2023does}, especially in tasks that require long-range interactions in order to be solved.

Graph rewiring is a technique which modifies the edges within the original graph to facilitate the exchange of messages between distant nodes and counteract the over-squashing issue. This approach has gained attention in recent research efforts \cite{bruelgabrielsson2022rewiringwithpe, abboud2022shortest, bodnar2021cw, bodnar2021simplicial, arnaizrodriguez2022diffwire, karhadkar2023fosr}. More recently, researchers have explored the use of expander graphs as a solution to address the over-squashing problem \cite{deac2022expander, banerjee2022oversquashing}. In this approach, message passing takes place alternately on the original graph and an expander graph. Expander graphs offer several advantageous properties for information propagation, such as sparsity, high connectivity, and logarithmic diameters, which enable efficient signal propagation across the graphs with minimal message passing steps.

On the other hand, hypergraphs provide a representation for capturing higher-order interactions within complex data \cite{feng2018higher, aktas2021higher, zhou2006hypergraphs}. While previous work \cite{deac2022expander} has utilised expander graphs in the context of pair-wise interactions, we aim to investigate the potential benefits of introducing higher-order interactions within expander graphs.


Our approach involves exchanging messages on bipartite expanders that represent hypergraphs. We construct random bipartite expanders using two algorithms: one based on perfect matchings, and the other using Ramanujan graphs. We demonstrate the effectiveness of our models by evaluating them on both synthetic and real-world data.

\begin{figure}
    \centering
    \includegraphics[width=\textwidth]{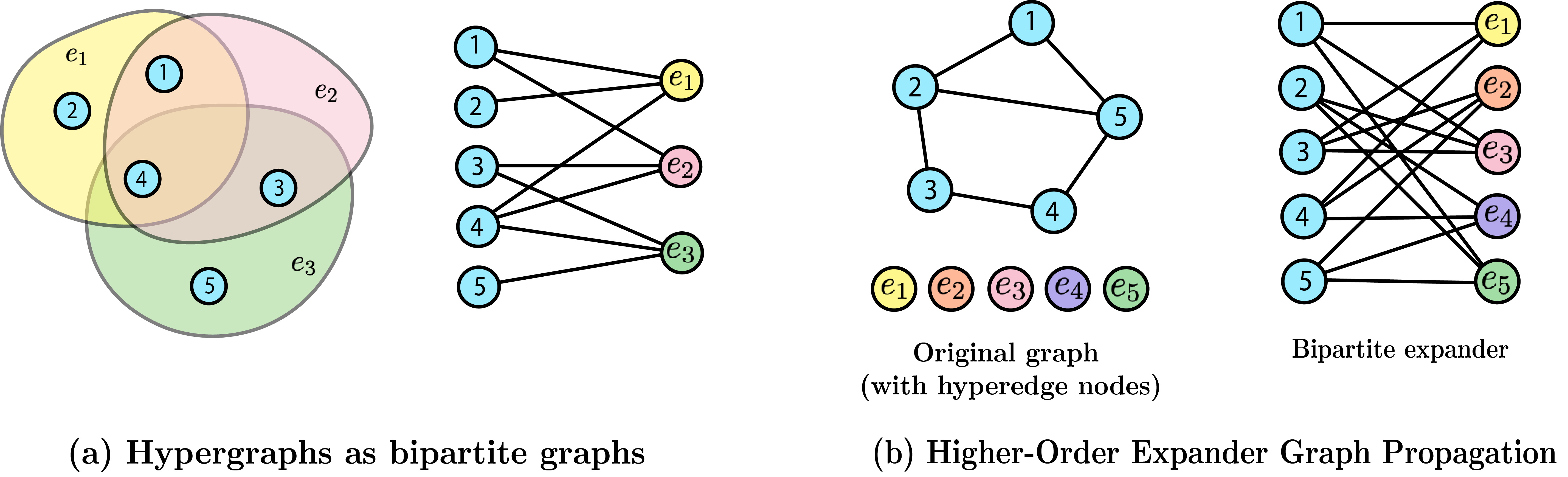}
    \caption{(a) Given a hypergraph with edges between sets of nodes of arbitrary cardinality, we can construct a corresponding bipartite representation, with one set of nodes corresponding to the original nodes in the hypergraph, and the other set representing hyperedges. (b) We first augment the input graph with disconnected hyperedge nodes. Then, we construct bipartite expanders where input graph nodes are on the left-hand side, and hyperedge nodes on the right-hand side. We perform message passing on the original graph and bipartite expander alternately. Message passing on the bipartite expander first goes from graph nodes to hyperedge nodes, and then back to graph nodes.}
    \label{fig:framework-combined}
\end{figure}

\section{Related Work}
\label{sec:related}



\paragraph{Approaches to over-squashing} Methods for addressing over-squashing in graph neural networks fall into two categories: spatial and spectral \cite{digiovanni2023oversquashing}. Spatial methods focus on reducing the distance between distant nodes, achieved through methods like adding explicit edges \cite{bruelgabrielsson2022rewiringwithpe, abboud2022shortest}, leveraging higher-order structures \cite{bodnar2021cw, bodnar2021simplicial}, or reweighting edges with attention mechanisms \cite{muller2023attending}. On the other hand, spectral methods aim to increase the graph's Cheeger constant, measuring its ``bottleneckness", such as differentiable rewiring for Lovász bound optimization \cite{arnaizrodriguez2022diffwire}, first-order spectral rewiring for spectral gap optimization \cite{karhadkar2023fosr}, and the use of expander graphs \cite{deac2022expander, banerjee2022oversquashing}.



\paragraph{Expanders} Expander graphs have gained attention as a means to mitigate the over-squashing problem in graph neural networks, primarily due to their favorable spectral properties. In recent research, $4$-regular Cayley graphs have been utilised as templates for interleaving message passing on the input graph and the expander graph \cite{deac2022expander}. Concurrently, another approach utilises a random local edge flip algorithm based on expander graph construction \cite{banerjee2022oversquashing}. Our work builds upon this existing research by focusing on the capture of higher-order interactions between nodes using bipartite expander graphs.


\section{Theoretical Background}
\label{sec:background}

\subsection{Hypergraphs}

Standard graphs, $\mathcal{G} = (\mathcal{V}, \mathcal{E})$, consist of a set of \textit{vertices} $\mathcal{V}$ and pairwise \textit{edges} between them: $\mathcal{E} \subseteq \{(\textit{u}, \textit{v}) | (\textit{u}, \textit{v}) \in \mathcal{V}^2\}$ \cite{trudeau2013introduction}. Graphs are said to be \textit{k-regular} if every vertex is of degree $k$. Hypergraphs $\mathcal{H} = (\mathcal{V}, \mathcal{E})$ are a generalisation of graphs, whereby edges may occur between \textit{sets} of nodes, which may be of arbitrary size i.e. $\mathcal{E} \subseteq \{\mathcal{X} | \mathcal{X} \subseteq \mathcal{V}\}$ \cite{bretto2013hypergraph}. A hypergraph is said to be \textit{k-uniform} if each hyperedge is of cardinality $k$.  An example can be seen in Figure \ref{fig:framework-combined}(a).

Bipartite graphs, $\mathcal{B} = (\mathcal{L}, \mathcal{R}, \mathcal{E})$, are graphs whose vertices can be separated into two disjoint sets $\mathcal{L}$ and $\mathcal{R}$, with no edges between nodes within a single set. Hypergraphs may be represented as bipartite graphs: one set of nodes in the bipartite graph, say $\mathcal{L}$, corresponds to the original nodes in the hypergraph, and the other set of nodes in the bipartite graph, $\mathcal{R}$, represents the hyperedges. Edges in the bipartite graph $(\textit{l}, \textit{r})$ are formed if node $\textit{l}$ in the hypergraph belongs to hyperedge $r$. The bipartite graph representation is particularly useful in graph representation learning, as it enables one to utilise the array of tools developed for standard graphs on hypergraphs with minimal adaptation.

\subsection{Expander Graphs}
\label{ssec:expander_graphs}

Expander graphs have recently been identified as a promising approach to alleviate the over-squashing problem \cite{deac2022expander}. Expander graphs are \textit{sparse} graphs ($|\mathcal{E}| = O(|\mathcal{V}|)$) which have a low \textit{diameter}. Therefore, performing message passing on expander graphs enables messages to be propagated between any pair of nodes in a low number of hops, alleviating the issue of over-squashing.

We define an expander graph more formally as follows. Firstly, for any subset of vertices $\mathcal{A}$ from the graph $\mathcal{G} = (\mathcal{V}, \mathcal{E})$, the \textit{outer boundary} of $\mathcal{A}$, denoted $\partial_{\text{out}} (\mathcal{A})$, consists of the set of nodes adjacent to nodes in $\mathcal{A}$, which themselves don't belong to $\mathcal{A}$. Then, an expander graph is defined as follows \cite{murty04}:

\begin{definition}[Regular Expander Graphs]
\label{def:expander}

A $k$-regular graph $\mathcal{G} = (\mathcal{V}, \mathcal{E})$ is said to be a $c$-expander graph if

\begin{equation}
    \frac{|\partial_{\text{out}} (\mathcal{A})|}{|\mathcal{A}|} \geq c
\end{equation}

for all subsets $\mathcal{A} \subset \mathcal{V}$ with $|\mathcal{A}| \leq \frac{|\mathcal{V}|}{2}$.
\end{definition}

There are various known methods for constructing expander graphs (\cite{kowalski2019introduction}, \cite{davidoff2003elementary}). Several of these constructions are algebraic, and utilise Cayley graphs \cite{cayley1878desiderata} for deterministic construction. However, other construction methods rely on a family of graphs known as \textit{Ramanujan graphs}, whose members have spectral properties that make them excellent candidates for expander graphs.

For an undirected graph with $n$ vertices, it is known that its adjacency matrix has $n$ real-valued eigenvalues $\lambda_1 \geq \lambda_2 \geq \ldots \geq \lambda_n$. If the graph is $k$-regular, then the eigenvalues satisfy $k = \lambda_1 > \lambda_2 \geq \ldots \geq \lambda_n \geq -k$, and if the graph is also bipartite then $\lambda_n = -k$. Eigenvalues $\lambda_i \neq \pm k$ are referred to as \textit{non-trivial} eigenvalues. The largest magnitude non-trivial eigenvalue is denoted $\lambda(\mathcal{G}) = \max\limits_{|\lambda_i| < k} |\lambda_i|$, which leads to Ramanujan graphs being defined as follows \cite{murty2020ramanujan}:

\begin{definition}[Ramanujan Graphs]
\label{def:ramanujan}
    A $k$-regular graph $\mathcal{G}$ is said to be Ramanujan if it satisfies the property $\lambda(\mathcal{G}) \leq 2 \sqrt{k - 1}$.
\end{definition}

We shall see that this property leads to Ramanujan graphs having low diameters, which follows from several results connecting the spectral properties of graphs with their diameters. The first important result was derived by Chung \cite{chung1989diameters} (and was further refined by \cite{van1995eigenvalues}), which presents a bound on the diameter of graph $\mathcal{G}$ in terms of its largest magnitude non-trivial eigenvalue $\lambda(\mathcal{G})$:

\begin{theorem}
For a connected $k$-regular graph $\mathcal{G}$ with $n$ vertices, its diameter is bounded by:
\begin{equation}
    \alpha + \frac{\log \left(\frac{2n}{\alpha}\right)}{log\left(\frac{k + \sqrt{k^2 - \lambda(\mathcal{G})^2}}{\lambda(\mathcal{G})}\right)}
\end{equation}

with $\alpha = 2$ in the case of bipartite graphs and $1$ otherwise. 
\end{theorem}

Hence, it follows that in order to minimise the diameter of the graph, we must minimise $\lambda(\mathcal{G})$. The Alon-Boppana bound \cite{nilli1991second} gives an asymptotic lower bound on $\lambda(\mathcal{G})$:

\begin{theorem}[Alon-Boppana Bound]
All sufficiently large $k$-regular graphs $\mathcal{G}$ satisfy:
\begin{equation}
    \lambda(\mathcal{G}) \geq 2 \sqrt{k - 1} - o(1)
\end{equation}    

with the asymptotic behaviour in the $o(1)$ term coming from the number of nodes in the graph $n$ going to infinity: $n \to \infty$.
\end{theorem}

Therefore, asymptotically, Ramanujan graphs have the smallest possible value of $\lambda(\mathcal{G})$, and so have asymptotically minimal diameters. This property can further be linked with the \textit{expander constant} of the resulting graph, $c$, defined in Definition \ref{def:expander}. It has been shown \cite{murty04} that:

\begin{equation}
    \frac{|\partial_{\text{out}} (\mathcal{A})|}{|\mathcal{A}|} \geq (k - \lambda_2(\mathcal{G})) \frac{|\mathcal{V} \setminus \mathcal{A}|}{|\mathcal{V}|}
\end{equation}

It follows from Definition \ref{def:expander} that the resulting expander constant is $(k - \lambda_2(\mathcal{G}))/2$. In order to maximise this, and hence generate expander graphs with good expansion properties, it follows that we should minimise $\lambda_2(\mathcal{G})$. Since we have $\lambda_2(\mathcal{G}) \leq \lambda(\mathcal{G})$, Ramanujan graphs also asymptotically minimise $\lambda_2(\mathcal{G})$. Therefore, it is clear that Ramanujan graphs are excellent expanders with low diameters and high expander constants.

Furthermore, even $k$-regular graphs make good expanders. An alternative definition of expander graphs is given in terms of their \textit{edge expansion}, $h(\mathcal{G})$. Given a subset of vertices $\mathcal{A} \subseteq \mathcal{V}$, the \textit{edge boundary} of $\mathcal{A}$, denoted $\partial \mathcal{A}$, is the set of edges with one end in $\mathcal{A}$, and the other end outside of $\mathcal{A}$. Then, an alternative definition of an expander graph is the following \cite{murty2020ramanujan}:

\begin{definition}[Regular Expander Graphs - Edge Expansion Definition]
A $k$-regular graph of $n$ vertices is an $(n, k, \delta)$-expander if its \textit{edge expansion}, $h(\mathcal{G})$ satisfies the following inequality:

\begin{equation}
h(\mathcal{G}) = \min\limits_{\mathcal{A} \subset \mathcal{V}: |\mathcal{A}| \leq n/2} \frac{|\partial \mathcal{A}|}{|\mathcal{A}|} \geq \delta     
\end{equation}
\end{definition}

A theorem from Dodziuk \cite{dodziuk1984difference} gives us a lower bound on $\delta$ for $k$-regular graphs:

\begin{theorem}
If graph $\mathcal{G}$ is $k$-regular then:

\begin{equation}
    \frac{k - \lambda(\mathcal{G})}{2} \leq h(\mathcal{G}) \leq \sqrt{2k(k-\lambda(\mathcal{G}))}
\end{equation}
\end{theorem}

\section{Higher-Order Expander Graph Propagation}
\label{sec:high-order-egp}

\subsection{Bipartite Expanders}
\label{sec:bipartite-expanders}

We rely on the fact that hypergraphs can be represented as bipartite graphs, and construct \textit{bipartite expander graphs} to capture higher-order interactions and leverage expander graph properties.

\paragraph{Perfect matchings}

A \textit{matching} on a graph is defined as a set of edges without common vertices \cite{cormen2022introduction}, and a \textit{perfect matching} is a matching which contains all vertices of the graph. One approach to constructing bipartite expander graphs is taking the union of $k$ perfect matchings \cite{odonnell2013}. By ensuring that the $k$ matchings are disjoint (i.e. contain no common edges), we guarantee that the resulting bipartite graph will be $k$-regular. As detailed in section \ref{ssec:expander_graphs}, $k$-regular graphs have good expansion properties, and so $k$-regular random bipartite graphs are good candidates for expander graphs. 

\paragraph{Random Ramanujan bipartite graphs}
In addition to generating random $k$-regular graphs, due to the utility of Ramanujan graphs as expanders, we also added the ability to check that the resulting graphs satisfied the Ramanujan property, given in definition \ref{def:ramanujan}. This incurs $O(|\mathcal{V}|^3)$ time complexity due to the need to calculate the eigenvalues of the graph's adjacency matrix.

\subsection{Framework}

As illustrated in Figure \ref{fig:framework-combined}(b), given an input graph $\mathcal{G}=(\mathcal{V}, \mathcal{E})$, we first augment it with $|\mathcal{V}|$ hyperedge nodes, call it the set $\mathcal{H}$. We construct a bipartite expander $\mathcal{B}=(\mathcal{L},  \mathcal{R}, \mathcal{E})$ where $\mathcal{L}=\mathcal{V}$ and $\mathcal{R}=\mathcal{H}$. Next, we connect nodes between the two sides by constructing a set of edges $ \mathcal{E}$, either using perfect matchings or Ramanujan graphs as explained in section \ref{sec:bipartite-expanders}. In this way, we have a $k$-regular bipartite expander, where we set $k$ as a hyperparameter. 

To incorporate the expander graph into the model without losing the topology of the original graph, we follow \cite{deac2022expander} by interleaving message passing on the two graphs. Therefore, we perform message passing on the original graph in odd layers, and on the expander graph in even layers.

Lastly, we ignore all hyperedge node features when performing graph pooling, which means only node representations from the input graph are used to predict the final graph classification. 

\subsection{Message passing on bipartite expanders} 

 We perform message passing in two directions sequentially, first from original graph nodes to hyperedge nodes, and then back, as shown in Figure \ref{fig:framework-combined}(b). This allows each hyperedge node to serve as a communication hub for $k$ graph nodes, enabling \textit{higher-order} message passing which goes beyond pair-wise interactions. In practice, the hyperedge node features are initialised with 0s. During message passing, we experiment with two handling methods. One allows hyperedge node features to be learnt end-to-end, whilst the other simply aggregates the messages from the original graph nodes at the hyperedge nodes via summation followed by a linear layer.

\section{Evaluation}
\label{sec:evaluation}

\subsection{Models} 

\paragraph{GIN}

We used Graph Isomorphism Network (GIN) \cite{xu2018how} as a baseline. Given $h_v^{(l)}$ as the representation for node $v$ at layer $l$, $\mathcal{N}(v)$ as the neighbours for node $v$, and $\epsilon$ as a learnable parameter, a GIN convolutional layer can be formulated as $h_v^{(l)} = \text{MLP}^{(l)} \left ( (1+\epsilon^{(l)})  \cdot h_v^{(l-1)}  + \sum_{u \in \mathcal{N}(v)} h_u^{(l-1)} \right )$. We perform graph-level pooling by averaging the node representations after the final GIN layer, and then use a task-specific loss function to evaluate the graph classification results.

\paragraph{Our models} We use GIN to perform message-passing on the augmented bipartite expander graph. We experiment with two bipartite expander construction methods, one based on perfect matchings (\textbf{GIN+PM}), and the other one additionally imposes the Ramanujan condition (\textbf{GIN+RM}). Furthermore, we test two ways of handling the arbitrary hyperedge node features, depending on whether they are learned (\textbf{learned features}) or aggregated via summation (\textbf{summation}).

\subsection{Set-up} 

We follow the same hyperparameter set-up as \cite{hu2020ogb} for graph-property prediction tasks. For a fair comparison, we use the same number of layers for all models. Note that we treat message passing in two directions on the bipartite graph as one layer, because in principle nodes in the original graph only get updated once. 

\section{Results}
\label{sec:results}

\subsection{Tree-NeighborsMatch}

\begin{figure}
\begin{floatrow}
\ffigbox{%
   \includegraphics[width=0.4\textwidth]{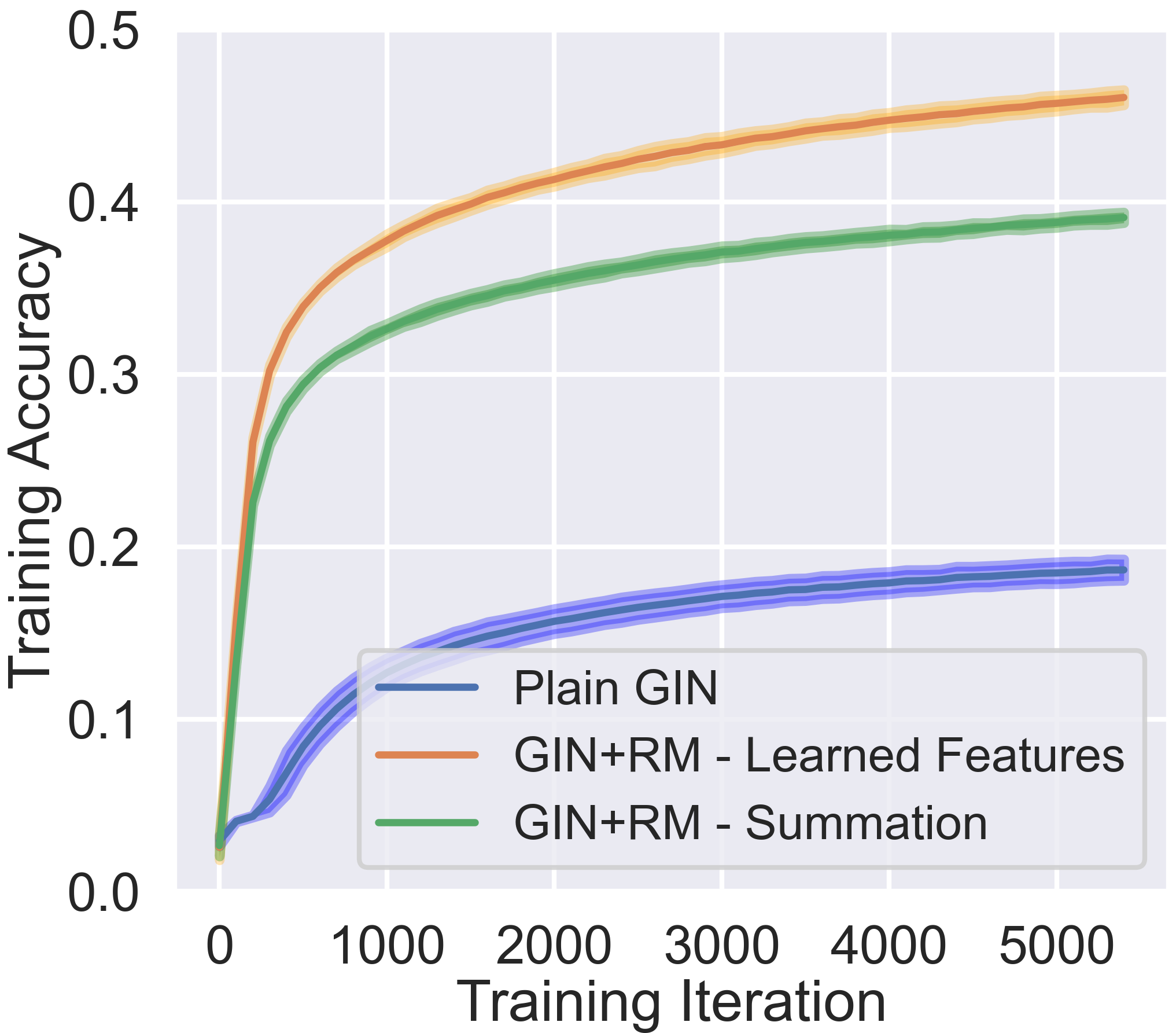}
}{
  \caption{Mean training accuracy ($\pm$ STD) on the Tree-NeighborsMatch dataset \cite{alon2020oversquashing} with binary trees of \texttt{depth=5}. It is clear that the interleaving message passing on the original graph and the higher-order expander graph helps mitigate the issue of over-squashing.
  }\label{fig:tree_neighbors}
}
\capbtabbox{%
  \footnotesize
  \begin{tabular}{c|c}
         \toprule 
          \textbf{Model} & \textbf{Test ROC-AUC} \\
         \midrule  
         Plain GIN \cite{hu2020ogb} & $0.7558 \pm 0.0140$ \\
         EGP \cite{deac2022expander} & $\color{Red}{0.7934 \pm 0.0035}$  \\
         \midrule
         GIN+PM+Learned Features &  \color{Dandelion}{$0.7742 \pm 0.0104$} \\
         GIN+PM+Summation & \color{Green}{$0.7751 \pm 0.0138$}  \\ 
         GIN+RM+Learned Features &  $0.7628 \pm 0.0132$  \\ 
         GIN+RM+Summation & $0.7737 \pm 0.0138$  \\ 
         \bottomrule 
    \end{tabular}
    \vspace{10mm}
}{
  \caption{Mean $\pm$ STD test ROC-AUC score on the \texttt{ogbg-molhiv} dataset \cite{hu2020ogb}, with various expander graph construction methods (PM, RM). \textcolor{Red}{Best}, \textcolor{Green}{Second Best} and \textcolor{Dandelion}{Third Best} results are coloured respectively.}\label{tab:molhiv}
}
\end{floatrow}
\end{figure}

We ran experiments on the synthetic Tree-NeighborsMatch dataset \cite{alon2020oversquashing} in the same manner as in the original expander graph propagation paper \cite{deac2022expander}. Namely, we ran the experiment on binary trees of \texttt{depth=5} with 6 GNN layers, at which depth standard GNN implementations begin to suffer from over-squashing. We ran experiments both without expander graphs (Plain GIN) and with a Ramanujan bipartite expander (GIN+RM), with two different methods of dealing with hyperedge node features. The regularity of bipartite expanders is set to $k=3$. Experiments were run $3$ times for $5400$ epochs. The results presented in Figure \ref{fig:tree_neighbors} demonstrate clearly that interleaving message passing on the original graph and the bipartite expander graph helps to mitigate the problem of over-squashing. 

\subsection{OGB - \texttt{molhiv}}

For real-world datasets, we first evaluated our models on the \texttt{ogbg-molhiv} dataset with $5$-regular bipartite expanders. As shown in Table \ref{tab:molhiv}, GIN+PM+Summation (bipartite expanders built with perfect matchings, using summation to handle hyperedge node features) gives the best performance. Interestingly, no strategies reach the same performance as the original expander graph paper \cite{deac2022expander}. We suggest that this may be because higher-order interactions are not useful in the \texttt{ogbg-molhiv} dataset, or it may be due to how we handled the learning of hyperedge node features. Investigating this would be useful in future work, using the Long Range Graph Benchmark \cite{dwivedi2022longrange}. 



\paragraph{Effect of Ramanujan condition} We observe in Table \ref{tab:molhiv} that Ramanujan expanders do not bring much benefit to the performances. This may suggest that enforcing the Ramanujan condition does not lead to a noticeable performance increase beyond merely using random $k$-regular bipartite expanders.

\subsection{OGB - \texttt{code2}}

\begin{table}[h]
    \centering
    \small
    \begin{tabular}{c|c}
        \toprule
         \textbf{Model} & \textbf{Test F$\mathbf{1}$ Score}  \\
         \midrule 
         Plain GIN \cite{hu2020ogb} & $\color{Dandelion}{0.1495 \pm 0.0023}$ \\
         EGP \cite{deac2022expander} & $\color{Green}{0.1497 \pm 0.0015}$ \\ 
         \midrule
         GIN + $3$-Regular Bipartite Expander + Learned Features & $\color{Red}{0.1519 \pm 0.0020}$ \\ 
         GIN + $3$-Regular Bipartite Expander + Summation & $0.1254 \pm 0.0029$ \\
         \bottomrule 
    \end{tabular}
    \caption{Mean $\pm$ STD test F1 scores on the \texttt{ogbg-code2} dataset. \textcolor{Red}{Best}, \textcolor{Green}{Second Best} and \textcolor{Dandelion}{Third Best} results are coloured respectively.}
    \label{tab:code2}
\end{table}

We also evaluated the performance on $\texttt{ogbg-code2}$. We generated random $3$-regular bipartite graphs using perfect matchings. As shown in Table \ref{tab:code2},  our best method, learning the features of the hyperedge nodes, outperforms both plain GIN \cite{hu2020ogb} and GIN + EGP \cite{deac2022expander}. One possibile explanation for this is that higher-order interactions may be helpful for this task, giving rise to a benefit from our approach.

\section{Conclusion}
\label{sec:conclusion}
We explored bipartite expander graphs as a solution for addressing over-squashing in graph neural networks. These hypergraphs are easily constructed, and even randomly generated $k$-regular bipartite graphs show favourable expansion properties. Our experiments yielded promising results, especially on Tree-NeighborsMatch and OGB - \verb|code2|, indicating that hypergraph expanders can help to mitigate over-squashing.

\begin{ack}
    We would like to thank Prof. Pietro Liò and Dr Petar Veličković for creating and delivering the ``L45: Representation Learning on Graphs and Networks'' course in Part III of the Computer Science Tripos at Cambridge, which made this paper possible.
\end{ack}



\bibliographystyle{unsrt}
\bibliography{ref}

\begin{thebibliography}{10}

\bibitem{kipf&welling2017}
Thomas~N. Kipf and Max Welling.
\newblock Semi-supervised classification with graph convolutional networks.
\newblock {\em CoRR}, abs/1609.02907, 2016.

\bibitem{velickovic2018}
Petar Veli{\v{c}}kovi{\'c}, Guillem Cucurull, Arantxa Casanova, Adriana Romero, Pietro Li{\`o}, and Yoshua Bengio.
\newblock Graph attention networks.
\newblock In {\em International Conference on Learning Representations}, 2018.

\bibitem{gilmer2017neural}
Justin Gilmer, Samuel~S. Schoenholz, Patrick~F. Riley, Oriol Vinyals, and George~E. Dahl.
\newblock Neural message passing for quantum chemistry.
\newblock In {\em Proceedings of the 34th International Conference on Machine Learning}, volume~70, pages 1263--1272. {PMLR}, 2017.

\bibitem{do2018chemical}
Kien Do, Truyen Tran, and Svetha Venkatesha.
\newblock Graph transformation policy network for chemical reaction prediction, 2018.

\bibitem{hamaguchi2017knowledge}
Takuo Hamaguchi, Hidekazu Oiwa, Masashi Shimbo, and Yuji Matsumoto.
\newblock Knowledge transfer for out-of-knowledge-base entities : A graph neural network approach.
\newblock In {\em Proceedings of the Twenty-Sixth International Joint Conference on Artificial Intelligence, {IJCAI-17}}, pages 1802--1808, 2017.

\bibitem{wu2019recommendation}
Qitian Wu, Hengrui Zhang, Xiaofeng Gao, Peng He, Paul Weng, Han Gao, and Guihai Chen.
\newblock Dual graph attention networks for deep latent representation of multifaceted social effects in recommender systems.
\newblock In {\em The World Wide Web Conference}. {ACM}, may 2019.

\bibitem{bronstein2021geometric}
Michael~M. Bronstein, Joan Bruna, Taco Cohen, and Petar Velickovic.
\newblock Geometric deep learning: Grids, groups, graphs, geodesics, and gauges.
\newblock {\em CoRR}, abs/2104.13478, 2021.

\bibitem{velickovic2022message}
Petar Veli{\v{c}}kovi{\'c}.
\newblock Message passing all the way up.
\newblock In {\em ICLR 2022 Workshop on Geometrical and Topological Representation Learning}, 2022.

\bibitem{xu2018how}
Keyulu Xu, Weihua Hu, Jure Leskovec, and Stefanie Jegelka.
\newblock How powerful are graph neural networks?
\newblock In {\em International Conference on Learning Representations}, 2019.

\bibitem{morris2018neural}
Christopher Morris, Martin Ritzert, Matthias Fey, William~L. Hamilton, Jan~Eric Lenssen, Gaurav Rattan, and Martin Grohe.
\newblock Weisfeiler and leman go neural: Higher-order graph neural networks, 2018.

\bibitem{oversmooth}
Qimai Li, Zhichao Han, and Xiao{-}Ming Wu.
\newblock Deeper insights into graph convolutional networks for semi-supervised learning.
\newblock In {\em Proceedings of the Thirty-Second {AAAI} Conference on Artificial Intelligence}, pages 3538--3545. {AAAI} Press, 2018.

\bibitem{alon2020oversquashing}
Uri Alon and Eran Yahav.
\newblock On the bottleneck of graph neural networks and its practical implications, 2020.

\bibitem{digiovanni2023does}
Francesco~Di Giovanni, T.~Konstantin Rusch, Michael~M. Bronstein, Andreea Deac, Marc Lackenby, Siddhartha Mishra, and Petar Veličković.
\newblock How does over-squashing affect the power of gnns?, 2023.

\bibitem{bruelgabrielsson2022rewiringwithpe}
Rickard Brüel-Gabrielsson, Mikhail Yurochkin, and Justin Solomon.
\newblock Rewiring with positional encodings for graph neural networks, 2022.

\bibitem{abboud2022shortest}
Ralph Abboud, Radoslav Dimitrov, and Ismail~Ilkan Ceylan.
\newblock Shortest path networks for graph property prediction.
\newblock In {\em The First Learning on Graphs Conference}, 2022.

\bibitem{bodnar2021cw}
Cristian Bodnar, Fabrizio Frasca, Nina Otter, Yu~Guang Wang, Pietro Liò, Guido Montúfar, and Michael Bronstein.
\newblock Weisfeiler and lehman go cellular: Cw networks, 2021.

\bibitem{bodnar2021simplicial}
Cristian Bodnar, Fabrizio Frasca, Yu~Guang Wang, Nina Otter, Guido Montúfar, Pietro Liò, and Michael Bronstein.
\newblock Weisfeiler and lehman go topological: Message passing simplicial networks, 2021.

\bibitem{arnaizrodriguez2022diffwire}
Adrian Arnaiz-Rodriguez, Ahmed Begga, Francisco Escolano, and Nuria Oliver.
\newblock Diffwire: Inductive graph rewiring via the lov\'asz bound, 2022.

\bibitem{karhadkar2023fosr}
Kedar Karhadkar, Pradeep~Kr. Banerjee, and Guido Montufar.
\newblock Fo{SR}: First-order spectral rewiring for addressing oversquashing in {GNN}s.
\newblock In {\em The Eleventh International Conference on Learning Representations}, 2023.

\bibitem{deac2022expander}
Andreea Deac, Marc Lackenby, and Petar Veličković.
\newblock Expander graph propagation, 2022.

\bibitem{banerjee2022oversquashing}
Pradeep~Kr. Banerjee, Kedar Karhadkar, Yu~Guang Wang, Uri Alon, and Guido Montúfar.
\newblock Oversquashing in gnns through the lens of information contraction and graph expansion, 2022.

\bibitem{feng2018higher}
Yifan Feng, Haoxuan You, Zizhao Zhang, Rongrong Ji, and Yue Gao.
\newblock Hypergraph neural networks.
\newblock {\em CoRR}, abs/1809.09401, 2018.

\bibitem{aktas2021higher}
Mehmet~Emin Aktas, Thu Nguyen, Sidra Jawaid, Rakin Riza, and Esra Akbas.
\newblock Identifying critical higher-order interactions in complex networks, 2021.

\bibitem{zhou2006hypergraphs}
Dengyong Zhou, Jiayuan Huang, and Bernhard Sch\"{o}lkopf.
\newblock Learning with hypergraphs: Clustering, classification, and embedding.
\newblock In B.~Sch\"{o}lkopf, J.~Platt, and T.~Hoffman, editors, {\em Advances in Neural Information Processing Systems}, volume~19. MIT Press, 2006.

\bibitem{digiovanni2023oversquashing}
Francesco~Di Giovanni, Lorenzo Giusti, Federico Barbero, Giulia Luise, Pietro Lio', and Michael Bronstein.
\newblock On over-squashing in message passing neural networks: The impact of width, depth, and topology, 2023.

\bibitem{muller2023attending}
Luis Müller, Mikhail Galkin, Christopher Morris, and Ladislav Rampášek.
\newblock Attending to graph transformers, 2023.

\bibitem{trudeau2013introduction}
Richard~J Trudeau.
\newblock {\em Introduction to graph theory}.
\newblock Courier Corporation, 2013.

\bibitem{bretto2013hypergraph}
Alain Bretto.
\newblock Hypergraph theory.
\newblock {\em An introduction. Mathematical Engineering. Cham: Springer}, 2013.

\bibitem{murty04}
Maruti~Ram Murty.
\newblock Ramanujan graphs.
\newblock {\em J. Ramanujan Math. Soc. 18, No.1 (2003) 1–20}.

\bibitem{kowalski2019introduction}
Emmanuel Kowalski.
\newblock {\em An introduction to expander graphs}.
\newblock Soci{\'e}t{\'e} math{\'e}matique de France, 2019.

\bibitem{davidoff2003elementary}
Giuliana~P Davidoff, Peter Sarnak, and Alain Valette.
\newblock {\em Elementary number theory, group theory, and Ramanujan graphs}, volume~55.
\newblock Cambridge university press Cambridge, 2003.

\bibitem{cayley1878desiderata}
Professor Cayley.
\newblock Desiderata and suggestions: No. 2. the theory of groups: graphical representation.
\newblock {\em American journal of mathematics}, 1(2):174--176, 1878.

\bibitem{murty2020ramanujan}
M~Ram Murty.
\newblock Ramanujan graphs: An introduction.
\newblock {\em Indian J. Discrete Math}, 6(2):91--127, 2020.

\bibitem{chung1989diameters}
Fan~RK Chung.
\newblock Diameters and eigenvalues.
\newblock {\em Journal of the American Mathematical Society}, 2(2):187--196, 1989.

\bibitem{van1995eigenvalues}
Edwin~R Van~Dam and Willem~H Haemers.
\newblock Eigenvalues and the diameter of graphs.
\newblock {\em Linear and Multilinear Algebra}, 39(1-2):33--44, 1995.

\bibitem{nilli1991second}
Alon Nilli.
\newblock On the second eigenvalue of a graph.
\newblock {\em Discrete Mathematics}, 91(2):207--210, 1991.

\bibitem{dodziuk1984difference}
Jozef Dodziuk.
\newblock Difference equations, isoperimetric inequality and transience of certain random walks.
\newblock {\em Transactions of the American Mathematical Society}, 284(2):787--794, 1984.

\bibitem{cormen2022introduction}
Thomas~H Cormen, Charles~E Leiserson, Ronald~L Rivest, and Clifford Stein.
\newblock {\em Introduction to algorithms}.
\newblock MIT press, 2022.

\bibitem{odonnell2013}
Ryan O’Donnell.
\newblock A theorist's toolkit lecture series, lecture 12, cmu 2013.
\newblock {\em https://www.cs.cmu.edu/~odonnell/toolkit13/lecture12.pdf}.

\bibitem{hu2020ogb}
Weihua Hu, Matthias Fey, Marinka Zitnik, Yuxiao Dong, Hongyu Ren, Bowen Liu, Michele Catasta, and Jure Leskovec.
\newblock Open graph benchmark: Datasets for machine learning on graphs.
\newblock In H.~Larochelle, M.~Ranzato, R.~Hadsell, M.F. Balcan, and H.~Lin, editors, {\em Advances in Neural Information Processing Systems}, volume~33, pages 22118--22133. Curran Associates, Inc., 2020.

\bibitem{dwivedi2022longrange}
Vijay~Prakash Dwivedi, Ladislav Ramp{\'a}{\v{s}}ek, Mikhail Galkin, Ali Parviz, Guy Wolf, Anh~Tuan Luu, and Dominique Beaini.
\newblock Long range graph benchmark.
\newblock In {\em Thirty-sixth Conference on Neural Information Processing Systems Datasets and Benchmarks Track}, 2022.

\end{thebibliography}


\end{document}